\documentclass[conference]{IEEEtran}
\IEEEoverridecommandlockouts

\makeatletter
\let\NAT@parse\undefined
\makeatother

\usepackage{amsmath,amssymb,amsfonts}
\usepackage{textcomp}
\usepackage{xcolor}
\usepackage{mathrsfs}
\usepackage{svg}
\usepackage{float}
\usepackage{graphicx}
\usepackage{comment}
\usepackage[absolute,overlay]{textpos}  
\usepackage{subcaption}
\usepackage{algpseudocode} 

\usepackage{booktabs} 
\usepackage{xcolor} 
\usepackage{pgfplots}
\pgfplotsset{compat=newest}

\usepackage[linesnumbered,ruled]{algorithm2e}
\def\BibTeX{{\rm B\kern-.05em{\sc i\kern-.025em b}\kern-.08em
    T\kern-.1667em\lower.7ex\hbox{E}\kern-.125emX}}
\def\BibTeX{{\rm B\kern-.05em{\sc i\kern-.025em b}\kern-.08em
    T\kern-.1667em\lower.7ex\hbox{E}\kern-.125emX}}

\graphicspath{{./figures/}}
\usepackage{amssymb}
\usepackage[numbers,sort&compress]{natbib}

\usepackage{amsthm}

\newcommand{\robots}{{\robotSet}}

\newcommand{\reward}{\mathbf{R}}

\newcommand{\robotSet}{\mathcal{R}}
\newcommand{\targetSet}{\mathcal{T}}

\newcommand{\constraintset}{\Omega}
\newcommand{\constraint}{\omega}
\newcommand{\checkConflict}{\Psi}
\newcommand{\pixelSet}{\mathcal{P\texttt{x}}}
\newcommand{\cost}{g}
\newcommand{\treeNode}{P}
\newcommand{\tree}{\texttt{TREE}}
\newcommand{\discountFactor}{\gamma}

\usepackage{balance} 
\usepackage[colorlinks=true, linkcolor=blue, urlcolor=blue, citecolor=blue]{hyperref}
\usepackage{cleveref}

\begin{document}

\title
{CoCap: Coordinated motion Capture for multi-actor scenes in outdoor environments\\
{\small \href{https://cocapture.github.io/}{\texttt{cocapture.github.io}}}%
}
\author{%
Aditya Rauniyar$^{1}$, Micah Corah$^{2}$, and Sebastian Scherer$^{1}$%

\thanks{$^{1}$A. Rauniyar, and S. Scherer are with the Robotics Institute, School of Computer Science at Carnegie Mellon University, Pittsburgh, PA, USA
{\tt\small \{rauniyar, basti\}@cmu.edu}}%
\thanks{$^{2}$M. Corah is with the Department of Computer Science at
  the Colorado School of Mines, Golden, CO, USA
  {\tt\small micah.corah@mines.edu}}
\thanks{This work is supported by the National Science Foundation under Grant No. 2024173 and supported by Defense Science and Technology Agency Singapore contract DST000EC124000205.}%
}

\maketitle

\begin{abstract}

Motion capture has become increasingly important, not only in computer animation but also in emerging fields like the virtual reality, bioinformatics, and humanoid training. Capturing outdoor environments offers extended horizon scenes but introduces challenges with occlusions and obstacles. Recent approaches using multi-drone systems to capture multiple actor scenes often fail to account for multi-view consistency and reasoning across cameras in cluttered environments. Coordinated motion Capture (CoCap), inspired by Conflict-Based Search (CBS), addresses this issue by coordinating view planning to ensure multi-view reasoning during conflicts. In scenarios with high occlusions and obstacles, where the likelihood of inter-robot collisions increases, CoCap demonstrates performance that approaches the ideal outcomes of unconstrained planning, outperforming existing sequential planning methods. Additionally, CoCap offers a single-robot view search approach for real-time applications in dense environments. 

\end{abstract}


\section{Introduction}

Unmanned Aerial Vehicles (UAVs) equipped with cameras have garnered considerable attention for their wide range of applications, including surveillance, search and rescue, environmental monitoring, mapping, inspection, and 3D reconstruction~\cite{gu_multiple_2018, alotaibi_lsar_2019, mccammon_ocean_2021, tong_integration_2015, kompis_informed_2021, ho20213d}.

\textbf{Motivation: }Recently, UAV teams have demonstrated effective view gathering for 4D pose reconstruction of single actors in outdoor environments~\cite{ho20213d}. This is particularly significant as traditional motion capture, typically confined to indoor studios, faces space limitations and incurs high costs for setting up the stage for stunts or other video shoots. Further research has extended this approach to multi-actor scenarios, where sequential view planning by a team of cameras ensures diverse and comprehensive coverage of a group of actors~\cite{suresh2024greedy}. Moreover, the development of a robust multi-camera aerial system still requires study.

\begin{figure}[t]
    \centering
    \includegraphics[width=\linewidth]{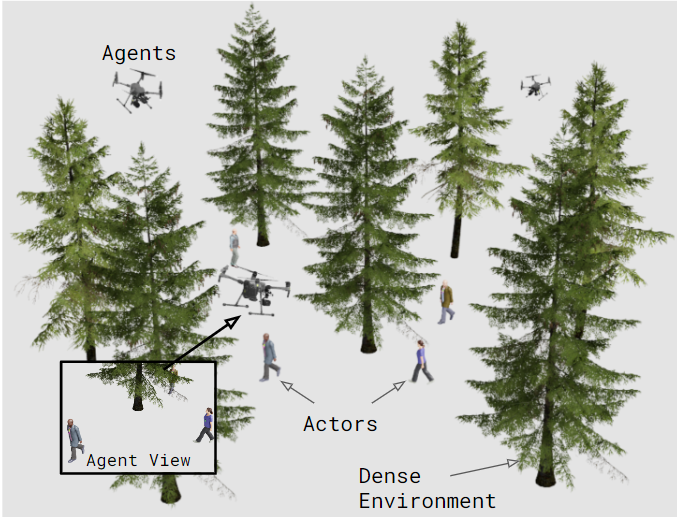}
    \caption{\textbf{Coordinated View Planning}: Coverage optimization on dynamic actors with flying cameras in an occlusion-aware and obstacle-clustered environment where camera extrinsic positions across robots are negotiated.}
    \label{fig:3Dscene}
\end{figure}

\textbf{Challenges: }Such motion capture applications in the wild present various challenges and requirements. First, since multi-view capture has been shown to improve 4D pose reconstruction for multiple actors \cite{liang2019shape, dong2019fast, zheng2023deep}, this demands a system of multiple camera-equipped drones capable of capturing diverse views of moving actors. Second, natural environments are filled with obstacles and occlusions, especially when cameras are maneuvered to obtain optimal coverage. This requires the drone system to not only manage obstacles and occlusions but also optimize for maximum coverage. Third, in these operational scenarios, as more camera-equipped drones are deployed, the system must also coordinate the views to avoid redundancy and prevent deadlocks in real-time.

\begin{figure*}[t]
  \centering
  \includegraphics[width=\linewidth]{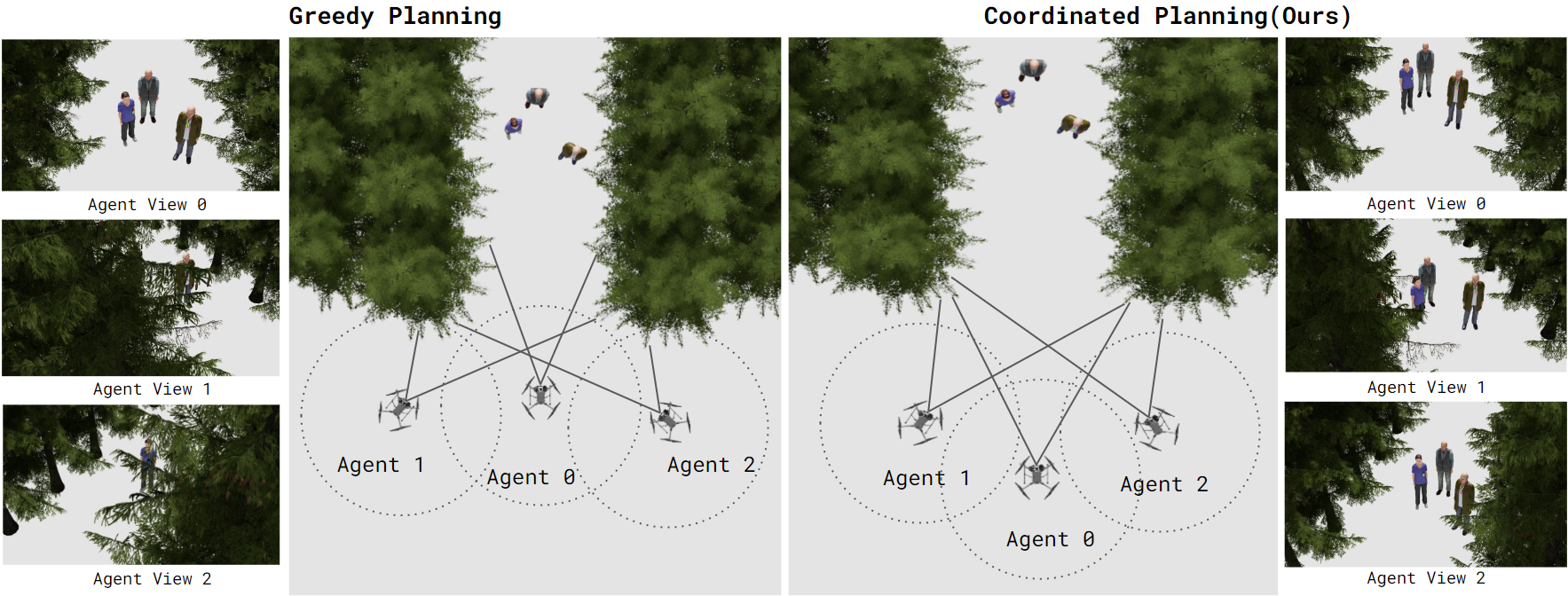}
   \caption{
        \textbf{Sequential (Greedy) View Planning}: On the left, there is the sequential view planning of multiple camera positions, where there are egocentric behaviors across multiple viewpoints as seen in the three camera outputs on the left under greedy planning. 
        \textbf{Coordinated planning}, on right: we propose a coordinated view planning approach where there is pixel-level negotiation amongst view positions to allow non-egocentric behaviors as seen in the three camera outputs on the right under coordinated planning.
        In general, we refer to behaviors where agents prefer to optimize their own reward to potential detriment of others as \emph{egocentric} and behaviors where agents act with respect for mutual constraints \emph{non-egocentric}.
    }
  \label{fig:CoCap}
\end{figure*}

\textbf{CoCap: }
To address these challenges, we develop an efficient method for incorporating constraints into the system as more drones are deployed. Drawing inspiration from Conflict-Based Search (CBS) \cite{sharon_conflict-based_2015}, we introduce constraints to the agents (drones) as conflicts arise. We define conflicts as instances where two agents enter a collision state, and we resolve this by applying constraints to one of the agents while maintaining a constraint tree. Our proposed approach of Coordinated motion Capture (CoCap) is tested in simulation, utilizing the reward structure employed by GreedyPerspectives \cite{suresh2024greedy}. Additionally, for real-time applications, we propose a single-agent view search method that prioritizes actor coverage and uses a heuristic to explore near-optimal viewpoints. More details on the simulation setup and reward structure can be found in our earlier research with GreedyPrespectives \cite{suresh2024greedy}.

\textbf{Contributions: } Main contributions are as follows:
\begin{itemize}
    \item A multi-camera view planning system that promotes diverse coverage from multiple viewpoints while coordinating robots to balance inter-robot conflicts with individual view rewards.
    \item Scenarios of multi-actor obstacle clustered environments that are particularly challenging for multi-camera equipped drones.  
    \item Heuristic-based single agent view planning framework for real-time applications. 
\end{itemize}


\section{Related Work}

\textbf{Motion capture of actors in natural settings}: Recently, there has been growing interest in developing motion capture systems using aerial cameras, particularly for outdoor environments. AirCapRL \cite{tallamraju_aircaprl_2020} introduces a deep reinforcement learning approach that learns an optimal policy for camera formations, focusing on achieving ideal viewing angles. However, this method struggles with reasoning about obstacles and occlusions in the environment and does not address scenarios involving multiple actors. Ho et al. \cite{ho20213d} tackled the first limitation by developing a formation planner constrained to optimal views around an actor using a spherical grid, though it still does not account for multiple actors. One approach that addresses the multiple actor scenario is by Hughes et al. \cite{hughes2024cdc}, who maximize Pixels-Per-Area (PPA) over multiple actors represented as polygonal cylinders. The GreedyPerspectives \cite{suresh2024greedy} further improves this method by addressing obstacles and occlusions. However, although these approaches can handle multiple actors, they are not intended for crowded and obstacle-dense environments where view rewards and inter-robot constraints may conflict. CoCap addresses these challenges by introducing system constraints as conflicts arise and employing a heuristic-driven view planning method, guided by rewards from maximizing coverage across multiple actors. 

\textbf{Collision-Free Navigation for Multi-Robot Teams}: Designing dense multi-robot systems presents significant challenges, particularly in conflict resolution. One widely used approach is prioritized planning, where robots are assigned a fixed priority sequence and planned sequentially based on their priority ID. This method has been employed in multi-actor view planning, such as in GreedyPerspectives \cite{suresh2024greedy}. However, prioritized planning can produce suboptimal in constrained and conflict-rich scenarios, as the preset order can cause later robots to encounter an increasing number of constraints, potentially resulting in deadlocks or planning failures. A more efficient approach to constraint development is Conflict-Based Search (CBS) \cite{sharon_conflict-based_2015}, where constraints are dynamically introduced as conflicts arise during the expansion of a constraint tree. In multi-robot systems, conflicts typically occur when robots attempt to occupy the same position at the same timestep. Despite the success of CBS in navigation, its application in perception planning systems remains under-explored. CoCap extends the CBS approach to motion capture in multi-actor scenarios, specifically in outdoor environments with complex obstacles and occlusions.


\section{Problem Formulation}

\begin{figure}[ht]
\centering
    \includegraphics[width=\linewidth]{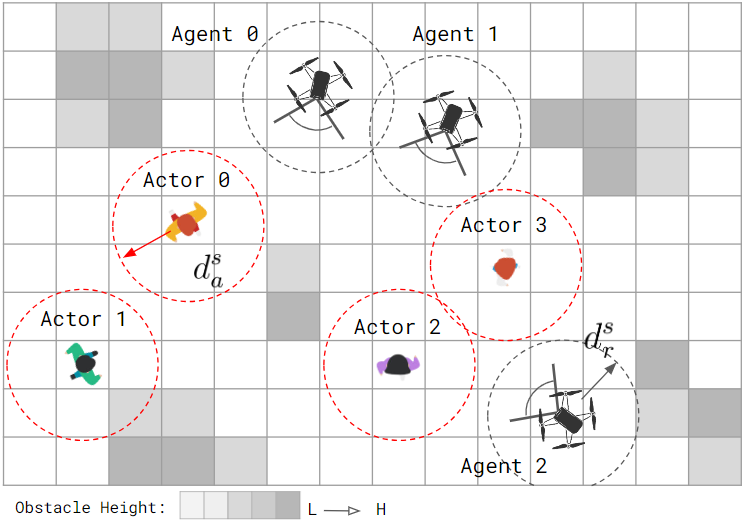}
    \caption{Problem representation of the gimbaled camera (also formulated as a robot, $\robots$) with projection matrix facilitating coverage on dynamic targets with occlusion and obstacles.}
    \label{fig:enter-label}
\end{figure}

We consider a multi-robot system equipped with gimbaled cameras that aims to monitor dynamic actors in an environment with obstacles. Each robot moves within a finite graph, selecting actions based on rewards tied to visual coverage. The following notation formalizes the setup, including robots, actors, and their observed faces, along with the workspace and reward structure.

Consider a set of robots $\robotSet=\{1,\ldots,N_r\}$, and a set of actors $\mathcal{A}=\{1,\ldots,N_a\}$ each with a set of faces $\mathcal{F}_j = \{1,\ldots,N_{j,f}\}$ where $j \in \mathcal{A}$. Also, let all the sets of faces across all the actors be  $\mathcal{F} = \{\mathcal{F}_1, ..., \mathcal{F}_{N_a}\}$. Each of the faces, $\mathcal{F}_j: j \in N_a$, are associated with a set of pixel coverage by agents, $\pixelSet = \{\pixelSet_{111}, ..., \pixelSet_{ijk}\}$, where $i \in N_r$, $j \in N_a$, and $k \in N_{j,f}$. All the robots move in a workspace represented as a finite graph $\mathcal{G} = (\mathcal{V}, \mathcal{E})$, and share a global clock that start at $t = 0$. Vertex set $\mathcal{V}$ represents the set of all the possible locations of $i^{th}$ robot at time $t$ as $v^{i}_t \in \mathcal{V}$ and the edge set $\mathcal{E}$ as actions $e^{i}_t \in \mathcal{E}$ of $i^{th}$ robot at time $t \in \{0,\ldots,T\}$ where $i \in \robotSet$. The reward for each edge is a $M$-dimensional nonnegative vector reward($e^{i}_t$) $\in \mathbb{R^+}^M \setminus \{0\}$.

Let $\xi^i(v_1^i, v_l^i)$ be a path that connects the vertices $v_1^i$ and $v_l^i$ via a sequence of vertices $(v_1^i, v_2^i, ..., v_l^i) \in \mathcal{G}$. Let $g^{i}(\xi^i(v_1^i, v_l^i))$ denote the $M$-dimensional reward vector associated with the path $\xi^i(v_1^i, v_l^i)$ as the sum of all the reward vectors of all the edges present in the path, $i.e.,  g^{i}(\xi^i(v_1^i, v_l^i)) = \sum_{j = 1, ..., l-1}  (v_j^i, v_{j+1}^i)$, in short, $\cost = J(\Xi)$.

Let $v^{i}_o, v^i_f \in \mathcal{V}$ represent the initial location and destination of robot $i$ respectively. For simplicity, we denote a path from $v^{i}_o$ to  $ v^i_f$ for robot $i$ as $\xi^i$. A joint path for all robots, denoted as $\xi = (\xi^1, \xi^2, ..., \xi^{N_r})$, is referred to as a solution.

\textbf {Objective: } We employ submodular optimization techniques to synchronize the actions of a multi-robot team and to study the associated objective. The observation of each actor $j$ is quantified in terms of the pixel density ($\frac{px}{m^2}$) obtained from a linear camera model's image. We define two functions, $\texttt{cov}(x_{i,t}, j, f) \to \mathbb{R}$, returning the pixel density for a specific actor's face observed from a robot's position, and $\texttt{covp}(j,f) \to \mathbb{R}$, returning the cumulative pixel density from all past observations. This enables us to express the incremental coverage gain:

\begin{equation}
  \begin{aligned}
\texttt{covm}(x_{i,t}) & = \sum_{j \in \mathcal{T}}^{} \sum_{f \in \mathcal{F}_j}^{} \sqrt{\texttt{cov}(x_{i,t},j,f) + \texttt{covp}(j,f)} \\
& - \sqrt{\texttt{covp}(j,f)}
  \end{aligned}
  \label{eq:coverage_metric}
\end{equation}

\begin{algorithm}[ht]
\DontPrintSemicolon
\caption{Coordinated View Planner}\label{alg:BB_MOVP}
\KwData{$X_{\texttt{init}}$, $\mathcal{T}$ trajectories , envHeightMap, agentMaxMotion}
\KwResult{$U^\dag$}
Initialization() \;
$\tree \gets GreedyPlanningWithoutConstraints()$\label{line:greedy}\;
Initialize $U_{\text{seq}} \gets \{\}$\;

\While{$\tree$ not empty}{
    $P_k \gets (\xi_k, \constraintset_k, \pixelSet_k, \cost_k)$ // \tree.pop()\;
    \If{no conflict detected in $\xi_k$}{
        return $P_k$
    }
    $\constraintset \gets$ Split detected conflict\;
    \ForAll{$\constraint_i \in \constraintset$}{
        $\constraintset_{li} \gets \constraintset_k \cup \{\constraint_i\}$\;
        $\pixelSet_l \gets \pixelSet \setminus \{\pixelSet_{itk}\}$\;
        $\xi^{i*} \gets$ LowLevelViewSearch($i, \constraintset_l, \pixelSet_l$)\;
        $\pixelSet_l^* \gets \pixelSet_l \cup GetCoverage(\xi^{i*})$\;
        $\xi_l^* \gets \xi_l \setminus \{\xi^i\} \cup \{\xi^{i*}\}$\;
        $g_l \gets GetObjectiveValue(\xi_l^*, \pixelSet_l^*)$\;
        $P_l \gets \{\xi_{l}^*, \constraintset_l, \pixelSet_{l}^*, \cost_l\}$\;
        $\tree \gets P_l$

    }
}

return $U_{\text{seq}}$\;
\end{algorithm}


\section{Brief Overview on Conflict-Based Search}

Conflict-Based Search \cite{sharon_conflict-based_2015} is a two-level search that creates a binary tree to resolve conflicts in the agent paths on a high level and runs an optimal low-level search algorithm for a Multi-Agent Path Finding (MAPF) problem.

\textbf{Conflict Resolution}: Consider a pair of agents, $i$ and $j$, each following their respective paths $\xi^i$ and $\xi^j$. To detect conflicts between these paths, we utilize the function $\checkConflict(\xi^i, \xi^j)$. This function returns either an empty set if no conflict is present, or it provides details about the first conflict encountered along the paths. A conflict occurs at time $t$ when agent $i$ is at position $v^i_t$ and agent $j$ is at position $v^j_t$. This conflict is denoted by $(i, j, v^i_t, v^j_t, t)$. To prevent such conflicts, a corresponding constraint is added to the path of either agent $i$ or agent $j$. This constraint is represented as $\constraint^i = (i, u^i_{a}, u^i_{b}, t)$, where $u^i_{a}$ and $u^i_{b}$ are selected from the set of possible locations $V$. This constraint is associated with agent $i$ and ensures that at time $t$, agent $i$ follows the specified path, thus avoiding the potential conflict.

Given a set of constraints $\constraintset$, let $\constraintset_i \subseteq \constraintset$ represent the subset of all constraints in $\constraintset$ that belong to agent $i$ (i.e., $\Omega = \bigcup_{i \in N_r} \Omega_i$). A path $\xi_i$ is consistent with respect to $\Omega$ if $\xi_i$ satisfies every constraint in $\Omega_i$. A joint path $\xi$ is consistent with respect to $\Omega$ if every individual path $\xi_i \in \xi$ is consistent.

\textbf{Two level Search}: In the high level while creating the constraint tree with each node containing $(\xi, \constraintset, g(\xi))$, each of them is in a priority queue. Initially, parent node $\pixelSet_0 \in \texttt{TREE}$ is computed for all the agents such that the constraint set, $\constraintset_o = \phi$, and $P_o = (\xi_0, \constraintset_o, g_o)$ gets pushed to \texttt{TREE}. Let, the total number of nodes created be denoted as $N_G$, and the total number of nodes expanded as $N_E: N_E \leq N_G$.

As each node, $k$, in the \texttt{TREE} gets explored based on the least $g$ value, $P_k = (\xi_k, \constraintset_k, g_k)$, $\checkConflict(\xi^i_k, \xi^j_k)$ gets called for all the sequence pairs $(i,j) \in N_r: (i \neq j)$. If there is no conflict detected, the solution, $\xi$, is found and the algorithm terminates. If there is a conflict detected, $(i, j, v^i_t, v^j_t, t)$, constraints to the agents are created in the following way. Each condition is created where constraint is added to either $i$ or $j$, such that $\constraint^i = (i, u^i_{a} = v^i_t, u^i_{b} = v^j_t, t)$ and $\constraint^j = (j, u^j_{a} = v^j_t, u^j_{b} = v^i_t, t)$. Each of these constraints leads to a new node in the \texttt{TREE}, where $P_{l^i} = (\xi_{l^i}, \constraintset_{l^i}, g_{l^i})$, and $P_{l^j} = (\xi_{l^j}, \constraintset_{l^j}, g_{l^j})$, such that $l^i \in N_G$ where new set of constraints is added to $i$ or $j$.  Here, $\constraintset_{l^i} = \constraintset_k \cup {\constraint^i}$, and $\constraintset_{l^j} = \constraintset_k \cup {\constraint^j}$. In each of these cases, for agent $i$, the algorithm updates the path $\xi_k^i$ in $\xi_{l^i}$ using the low-level search with a set of constraints $\constraintset_{l^i}$. A similar call is made for agent $j$ leading to $\xi_k^j$. If the low-level search is unable to find a solution for any of these cases, respective $P_{l^n}$ is discarded.

CBS solves a single objective cost function, $g$, optimally by iterative expansion of a node in the constraint tree based on the least $g$ val, resolving first agent-agent conflict (if exists) and creating a new node in the tree with new $\constraintset$, or returning the solution $P = (\xi, \constraintset, g)$.


\section{Coordinated View Planning}

Our proposed planner Alg.~\ref{alg:BB_MOVP} also works in a similar structure to CBS. With the following key differences:

\begin{itemize}
    \item \textbf{Greedy planning}: Rather than initializing with solutions to single-agent path-finding problems we initialize with an approximate solution to the joint multi-agent view planning problem that relaxes constraints between robots. 
    \item Tree Node (\treeNode$_{k}$): This node tracks pixel coverage by agents over actor faces, defined as $P = (\xi, \constraintset, \pixelSet, \cost)$, where $\xi$ is the joint robot path, $\constraintset$ the path constraints, $\pixelSet$ the total observed pixels, and $\cost$ the node's reward. 
    \item \textbf{Cost(\cost)}: The cost, \cost, is the overall reward of the tree node as denoted by  \cref{eq:coverage_metric}.  
    \item \textbf{Termination criteria}: Considering the computational expense, we return the first set of the solution $\treeNode$ without any conflicts. If there is no valid solution, a failure is reported.
    \item \textbf{Replanning}: As a new conflict between any two agents $(i,j)$ is detected using $\checkConflict$, the low level plans for agent $i$, with constraint set $\constraintset_{l^i}$  and $\pixelSet \setminus \{\pixelSet_{ijk}\}$ : $j \in N_a$, and $k \in N_{j,f}$. Similarly,  for agent $``j"$.
    \item \textbf{Low-level search}: Replace $A^*$ with a reward-guided solver using value iteration, focusing on coverage over actors and stable camera motion—key criteria for single-agent view planning.
\end{itemize}

\subsection{Constraint Tree Formation}
\textbf{Initialization}: During this state, with $(\constraintset_o = \phi, \pixelSet_o = \phi)$, as we plan for a series of agents, $i \in \robotSet$, we plan each agent greedily towards maximizing its reward using the single-agent view planner to produce the set of trajectories $\xi_o$. The root node, $\treeNode_o = \{\xi_o, \constraintset_o, \pixelSet_o, \cost_o\}$ gets added to \tree. As each agent, $j$, gets planned, the pixel context as passed such that $\pixelSet_j = \{\pixelSet_{jkl}\} : j < i, k \in \targetSet, l \in N_{k,f}.$

\textbf{Finding a solution}: The constraint tree, \(\tree\), is processed by removing its root node, represented as \(\treeNode_k = \{\xi_k, \constraintset_k, \pixelSet_k, \cost_k\}\). The removed node is then analyzed using \(\checkConflict\), which checks for conflicts between all pairs of agents \((i, j)\) where \(i, j \in N_r\) and \(i \neq j\). If no conflicts are detected, the current node represents a solution, and the high-level search with the $\tree$ terminates. While multiple conflict-free solutions may exist that maximize \(\cost\), the focus is on identifying the first solution for computational efficiency. The tree expansion is guided by the objective function, \(\cost = J(\Xi)\).

\textbf{Resolving conflicts}: If there exists a conflict between any two agents, $(i, j): i \neq j$, then agent $i$ is re-planned with constraint set $\constraintset_{l^i} = \constraintset_k \cup S^{ja}_t$ and $\pixelSet \setminus \{\pixelSet_{itk}\}$ : $t \in \targetSet$, and $k \in N_{t,f}$, and agent $j$ is re-planned with constraint set $\constraintset_{l^j} = \constraintset_k \cup S^{ia}_t$  and $\pixelSet \setminus \{\pixelSet_{jtk}\}$ : $t \in N_a$, and $k \in N_{t,f}$. The output trajectory of each agent $i$ from the low-level solver is updated to $(\xi^{i}_{l^i})$, and pixel coverage set to $\pixelSet_{l^i}$, such that $P_{l^i} = (\xi_{l^i}, \constraintset_{l^i}, \pixelSet_{l^i}, g_{l^i})$, where $g_{l^i}$ is calculated with the updated set of $\constraintset_{l^i}$ and $\pixelSet_{l^i}$. Similarly, for agent $``j"$, new constraints and pixel-set leads to $P_{l^j} = (\xi_{l^j}, \constraintset_{l^j}, \pixelSet_{l^j}, g_{l^j})$. $P_{l^i}$, and $P_{l^j}$ are added to the $\tree$ with their corresponding \cost, and the tree is rearranged such that the root node has max(\cost = $J(\Xi)$).

\begin{algorithm}[ht]
\DontPrintSemicolon
\caption{Single-Agent View Search}\label{alg:single-agent}
\KwData{$\xi^i_0$, $\mathcal{T}$ trajectories, envHeightMap($H$), agentMaxMotion($m^i$), $\pixelSet_i$, $\constraintset_i$}
\KwResult{$\xi^i$}
Initialize $\xi^i \gets \{\}$\;
$\tree \gets \{\reward_0, \xi^i_0, e^i_0, t \}$\;

\While{$\tree.\texttt{top}().t$ is not $T$}{
    $Q_k \gets (\overline{\reward_k}, \xi^i_k, e^i_{k-1}, k)$ \Comment*[r]{\tree.pop()}
    updateProcessedStates[$\xi^i_k$] = true\;
    updateStatesFromPrevAction($Q_k$)\;
    $e^i_{k_1}, ..., e^i_{k_{S^T}} \gets \texttt{availableActions}(\xi^i_k, k, \constraintset_i, H, m^i)$\;
    \ForAll{$e^i_{k_1}, ..., e^i_{k_{S^T}}$}{
        $\overline{\reward_{k+1}} \gets (\reward_{k+1} + \overline{\reward_k}) \discountFactor^{k-1}$\;
        \Comment*[r]{Cumulative reward $\times$ discount factor ($< 1$) for encouraging higher reward states in fewer timesteps}
        $Q_{k+1} \gets \{(\overline{\reward_{k+1}}, \xi^i_{k+1}, e^i_{k_n}, k+1)\}$\;
        \If{$processedStates[\xi^i_{k+1}] == \texttt{true} \,\textbf{or}\, \overline{\reward_{k+1}} < \tree[\xi^i_{k+1}].\reward$}{
            \texttt{continue}\;
        }
        $\tree \gets Q_{k+1}$\;
    }
}
$\xi^i \gets \texttt{BackTracking}(\texttt{processedStates})$\;

\Return $\xi^i$\;
\end{algorithm}

\subsection{Single-Agent View Planner}

The proposed single-agent view search draws from the $A*$ path planning algorithm, using a perceptual metric as a heuristic to guide node exploration, as shown in Algorithm \ref{alg:single-agent}. The algorithm begins at the camera-equipped UAV's start position, adding to $\tree$ all nodes with valid actions within the action space. Each node’s reward is adjusted by a discount factor to balance short-term and long-term rewards. A new $\treeNode_{k+1}$ is created, and if it’s either absent from $\tree$ or has a higher reward than an existing node, it’s added to $\tree$. Finally, the trajectory is computed by backtracking from the end position.

\begin{figure}[t]
    \centering
    \begin{subfigure}[b]{0.22\textwidth}
        \centering
        \includegraphics[width=\textwidth]{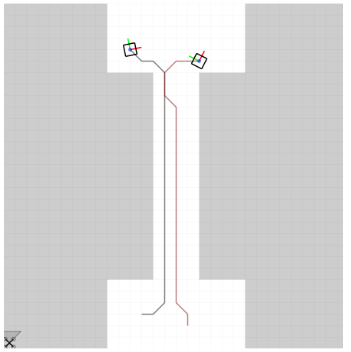}
        \caption{Corridor}
        \label{fig:corridor}
    \end{subfigure}
    \hfill
    \begin{subfigure}[b]{0.22\textwidth}
        \centering
        \includegraphics[width=\textwidth]{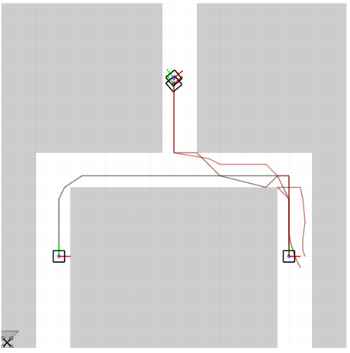}
        \caption{Bottleneck}
        \label{fig:bottleneck}
    \end{subfigure}
    \caption{Scenarios requiring high camera coordination due to conflicting trajectories. (a) Corridor scenario with two actors navigating narrow passageways. (b) Bottleneck scenario with four actors moving through a confined region with intersecting paths.}
    \label{fig:combined}
\end{figure}



\section{Experimental Results}

In this section, we present the experimental results of the multi-UAV camera system designed for dynamic actors. The simulations were conducted using a custom-built simulation environment to test the system under bottleneck and corrridor scenarios as shown in Fig. \ref{fig:combined}.

\subsection{Ego-Centric Test}

\begin{figure}[ht]
    \centering
    \begin{subfigure}[b]{0.45\textwidth}
        \includegraphics[width=\textwidth]{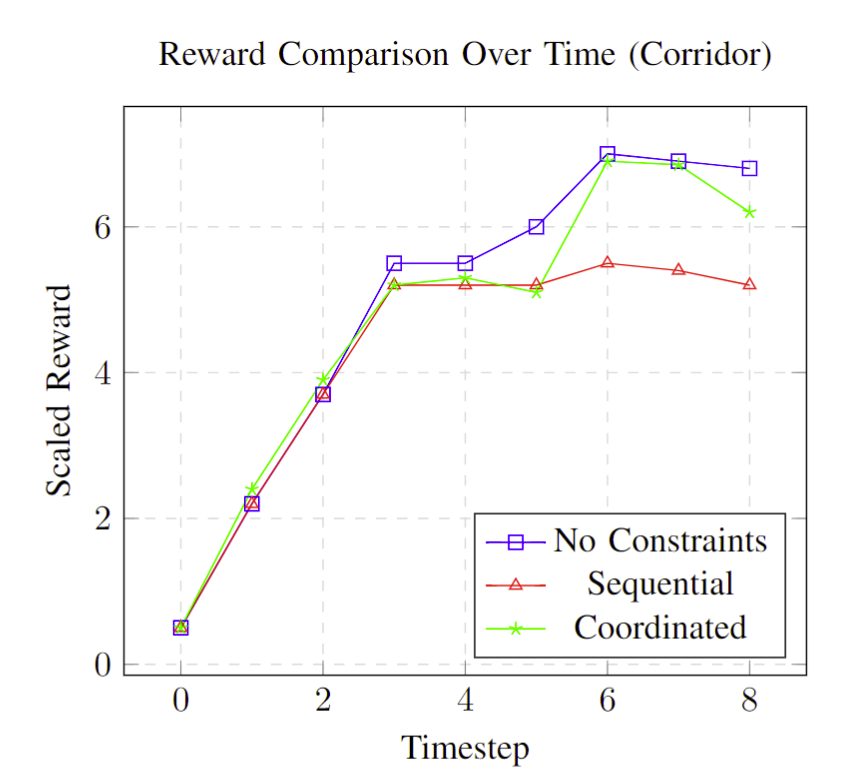}
        \caption{Reward comparison amongst different methods in corridor environment.}
        \label{fig:corridor_env}
    \end{subfigure}
    \hfill
    \begin{subfigure}[b]{0.45\textwidth}
        \includegraphics[width=\textwidth]{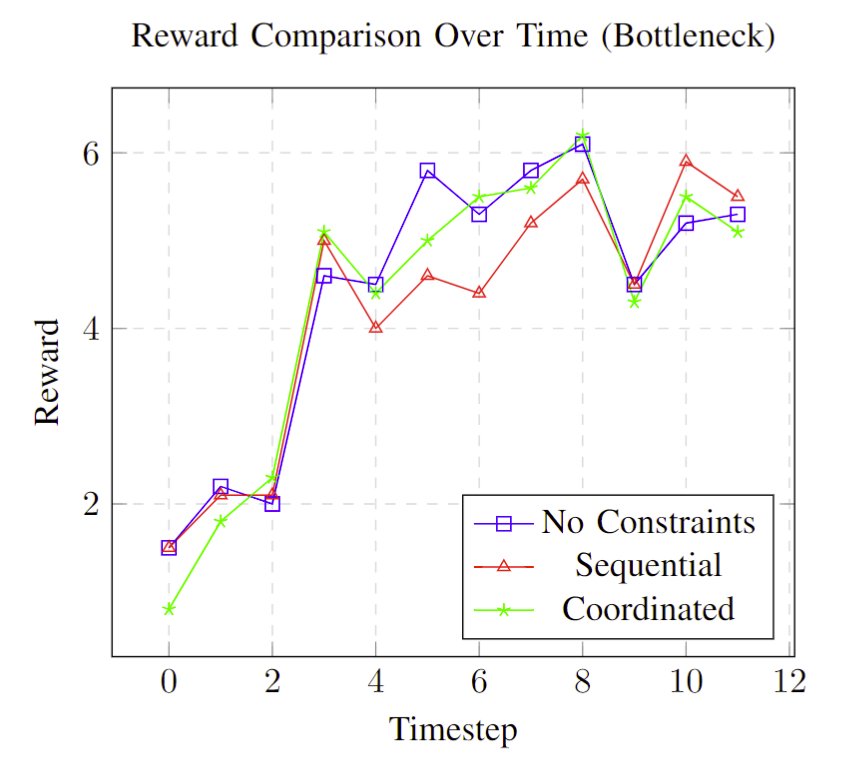}
        \caption{Reward comparison amongst different methods in bottleneck environment.}
        \label{fig:bottleneck_env}
    \end{subfigure}
    \caption{Scale rewards using multiple cameras performing view planning using No inter-robot Constraint, Sequential Constraint, and Conflict Based MDP Value Iteration, over total planning horizon for the environment.}
    \label{fig:reward_comparision}
\end{figure}

In this test, we evaluated scenarios where the agents exhibited non-egocentric behavior, such as one robot getting out of the way to allow another robot to pass as shown in \Cref{fig:CoCap}. The objective was to assess the system's ability to handle interactions and coordination among agents. We also evaluate the constraints developed in the system when comparing the no-inter-robot constraints, while adding inter-robot constraints, and while doing a conflict-based constraint assignment.

\Cref{fig:corridor_env} presents the scaled rewards of three view planning methods in a corridor environment with 8 timesteps, 2 robots, and 2 actors. The robots are randomly initialized outside the corridor while the actors move through it in the same direction. From timesteps 3 to 8, when the robots are inside the corridor, a clear gap emerges between the performance of the proposed sequential planning and planning without system constraints. The proposed Alg. \ref{alg:BB_MOVP} aims to narrow this gap, bringing performance closer to the unconstrained system.

Similarly, \Cref{fig:bottleneck_env} compares the three methods in a bottleneck environment over 11 timesteps with 4 robots and 4 actors, where actors from different corridors merge in the bottleneck area. From timesteps 3 to 9, the gap between the sequential planning view reward and planning without inter-robot constraints highlights the system's reward decrease as constraints are introduced. This indicates the need for an intelligent approach to adding constraints without significantly reducing the view reward. The proposed coordinated capture effectively bridges this gap, achieving performance comparable to the unconstrained method but with added inter-robot constraints.

Together, these two scenarios in \Cref{fig:reward_comparision}, with high obstacles and occlusions, demonstrate the proposed methods' effectiveness in maintaining high-view rewards despite added constraints. 

\subsection{Single-Agent Value Iteration vs. Search}

For offline planning approaches, the value iteration solver can compute sub-optimal view positions to achieve high view rewards as shown in GreedyPerspectives \cite{suresh2024greedy}. However, in more online settings, there is a need for more efficient algorithms. The proposed single-agent view search, described in Alg. \ref{alg:single-agent}, aims to address these use cases.

The Table \ref{tab:viewsearch_vs_valueIteration} compares the total reward (scaled) and compute time (in seconds) for different view planning methods in a corridor environment. The Sequential method with Value Iteration achieved a total reward of 4127 and a compute time of 482 seconds. The Coordinated method with Value Iteration yielded the highest total reward of 4662 but required significantly more compute time, at 3981 seconds. In contrast, the Coordinated method with View Search had a lower total reward of 3922 but was the fastest, with a compute time of only 2.7 seconds. This highlights a trade-off between achieving higher rewards and compute efficiency, showcasing the effectiveness of using single-agent view search in online settings.

\begin{table}
  \centering
\begin{tabular}{lcccc}
  \toprule
  \bfseries Method & \multicolumn{2}{c}{\bfseries Total Reward (scaled)} & \multicolumn{2}{c}{\bfseries Compute Time (s)} \\
  \cmidrule(r){2-3} \cmidrule(r){4-5}
                   & Corridor & Bottleneck & Corridor & Bottleneck \\
  \midrule
  Sequential (MDP) & 4127 & 5053 & 482 & -- \\
  CoCap (MDP) & \textbf{4662} & \textbf{5162} & 3981 & -- \\
  CoCap (Search) & 3922 & -- & \textbf{2.7} & -- \\
  \bottomrule
\end{tabular}
    \caption{Comparison of Total Reward and Compute Time across different methods for corridor and bottleneck scenarios. Missing compute time results for bottleneck are denoted as ``--".}
    \label{tab:viewsearch_vs_valueIteration}
\end{table}


\section{Conclusion and Future Work}
In conclusion, this research tackles the challenge of coordinated motion capture of multiple actors in complex, obstacle-dense environments using camera-equipped UAVs. We introduced a novel multi-robot coordinated view planning system, inspired by Conflict-Based Search (CBS) with an occlusion-aware objective. Evaluations in two scenarios demonstrate that our approach, CoCap, outperforms sequential planning, especially in narrow and obstacle-dense environments. Coordinated view planning closely matches the performance of a system without collision constraints, while outperforming sequential greedy planning. Additionally, our single-agent view search method provides a significant computational advantage over the single-agent value iteration solver currently being used.

While promising, the system has limitations. It relies on a 2.5D height map, limiting its handling of overhangs and coverage at doors and windows. Actors are oversimplified as cuboids, and real multi-UAV deployment hasn't been tested, potentially facing communication issues with the centralized planner. More testing is needed in denser, occluded environments, and the search heuristic lacks exploration strategies, especially when high rewards emerge later. Ensuring communication reliability is also crucial for real-world deployment.

\section*{Acknowledgment}

We would also like to thank Krishna Suresh, Yuechuan Hou, and Micah Nye for their assistance in developing parts of the work. Micah Corah primarily participated in this work while a Postdoctoral Fellow at CMU.  


\bibliographystyle{IEEEtranN}
{
  \small
  \bibliography{main}
}

\end{document}